%% file: main.tex
\documentclass[runningheads]{llncs}
\usepackage{hyperref}
\usepackage{graphicx}
\usepackage[acronym]{glossaries}
\usepackage{array}
\usepackage{multirow}
\usepackage{booktabs}
\usepackage{graphicx}
\usepackage{float}
\usepackage{url}

\newcommand{\fref}[1]{Figure~\ref{#1}}
\newcommand{\tref}[1]{Table~\ref{#1}}

\newcommand{\sref}[1]{Section~\ref{#1}}

\glsdisablehyper

\begin{document}
\title{Dataset and Benchmarking of Real-Time Embedded Object Detection for RoboCup SSL}
\titlerunning{Dataset and Benchmarking of Embedded Object Detection for SSL}
\author{Roberto Fernandes \and
Walber M. Rodrigues \and
Edna Barros}
\authorrunning{R. Fernandes et al.}
\institute{Centro de Informática, Universidade Federal de Pernambuco. Recife, Brazil.
\email{\{rcf6,wmr,ensb\}@cin.ufpe.br}\\
\url{robocin.com.br}}
\maketitle              % typeset the header of the contribution
%
\input{default_content/acronyms}

\input{default_content/abstract}
\input{chapters/introduction}
\input{chapters/related_work}
\input{chapters/dataset}
\input{chapters/evaluation}
\input{chapters/results}
\input{chapters/conclusion}
\input{default_content/acknowledgements}

\bibliographystyle{splncs04}
\bibliography{default_content/references}

\end{document}

%% file: default_content/acronyms.tex
% \newacronym{id}{abrev}{completo}

\newacronym{ssl}{SSL}{Small Size League}
\newacronym{msl}{MSL}{Middle Size League}
\newacronym{spl}{SPL}{Standard Platform League}
\newacronym{cnn}{CNNs}{Convolutional Neural Networks}
\newacronym{gpu}{GPUs}{Graphical Processing Units}
\newacronym{ap}{AP}{Average Precision}
\newacronym{ar}{AR}{Average Recall}
\newacronym{iou}{IoU}{Intersection over Union}
\newacronym{fps}{FPS}{Frames Per Second}
\newacronym{gan}{GAN}{Generative Adversarial Networks}

%% file: default_content/abstract.tex
\begin{abstract}

When producing a model to object detection in a specific context, the first obstacle is to have a dataset labeling the desired classes. In RoboCup, some leagues already have more than one dataset to train and evaluate a model. However, in the \gls{ssl}, there is not such dataset available yet. This paper presents an open-source dataset to be used as a benchmark for real-time object detection in \gls{ssl}. This work also presented a pipeline to train, deploy, and evaluate \gls{cnn} models in a low-power embedded system. This pipeline is used to evaluate the proposed dataset with state-of-art optimized models. In this dataset, the MobileNet SSD v1 achieves $44.88\%$ \acrshort{ap} ($68.81\%$ \acrshort{ap}$_{50}$) at 94 \gls{fps}, while running on an \gls{ssl} robot.

\keywords{Dataset \and Benchmark \and Deep Learning \and Object Detection.}
\end{abstract}

%% file: chapters/introduction.tex
\section{Introduction}
\label{introduction}

The \acrfull{ssl} is one of the most traditional leagues in the RoboCup. In this league, it is possible to precisely perform a wide range of dynamic plays every moment during a game. The decision-making process at each play needs to be fast to keep up with the fast-paced game, in which robots usually move at $3m/s$, and the ball reaches $6.5m/s$. These actions are possible due to the use of omnidirectional wheeled robots, and the use of SSL-Vision \cite{sslvision} as a global vision system. 

Due to the use of the SSL-Vision, all robots have all the field information, making it easy to design and develop a tactic. With this external vision system, a team in the \gls{ssl} is considered a semi-autonomous system. As a comparison, in \gls{msl} and \gls{spl} instead of using external information, each robot has its camera and vision system, limiting the information to which they have access. Thus, they are considered a fully autonomous system because each robot can perform a tactic without receiving external information.

A technical challenge \cite{tech_ssl} was introduced in 2019 to evolve the league, encouraging the teams to develop and propose a local vision system. This challenge aims to bring autonomy to an \gls{ssl} robot, in a similar way to \gls{msl} and \gls{spl}. A \gls{msl} robot fits in $52\times52\times80cm$, which can equip a  full-size computer, and a \gls{spl} robot can not be modified, since it uses the NAO as standard platform. However, an \gls{ssl} robot needs to fit in a cylinder with a height of 15cm and a diameter of 18cm \cite{rules_ssl}, which constrains the robot's vision system complexity.

In the first three steps of this challenge, a robot has to grab a stationary ball, find a goal, and score against a static defender robot without receiving any information from the SSL-Vision. Therefore, a robot has to detect a Robot, a Ball, and a Goal autonomously. It also has to respect the league requirements, except for the height restriction, creating a small room for hardware improvements.

The straightforward option to detect these objects uses scan lines and color segmentation to detect the ball \cite{tigers_scan}, as the league uses an orange golf ball. However, this approach can not detect robots and goals because they do not have a unique pattern. For instance, a team can use robots of any color, making it harder to use this technique. Besides, the color segmentation approach needs to be re-calibrated on each slight environment variation, as uneven illumination or field changes \cite{segmentation}.

The state-of-the-art of object detection relies on \acrfull{cnn} \cite{yolov4}, which given a labeled dataset, trains a model once, and does not need any other calibration or modification. Besides, this approach is robust to deal with occlusion, scale transformation, and background switch \cite{cnn_survey}, which makes \gls{cnn} strong candidates to use in the \gls{ssl}.

For other RoboCup's leagues, like \gls{spl} \cite{spl_dataset} and \gls{msl} \cite{dataset_msl}, there are public object detection datasets. Although, in the \gls{ssl} does not exist an open-source labeled dataset and creating a new one takes time, making the research and development of object detection models in this league even harder.

Therefore, given the \gls{ssl} technical challenge, the league constraints, and the lack of an open-sourced dataset, this paper has two main contributions:
\begin{itemize}
  \item Propose a novel open-source dataset for \gls{ssl}, containing labels for Robot, Ball, and Goal, intended to benchmark object detection in this league.
  \item Evaluate and compare \gls{cnn} models, respecting the league's hardware constraints while achieving an inference frequency of at least 24 \acrfull{fps}, real-time rate, necessary during actual games.
\end{itemize}

This paper's remainder is organized as follows: \sref{related_work} will present some related works. \sref{dataset} will detail the dataset. \sref{evaluation} will explain the evaluation methodology. \sref{results} will show and discuss the achieved results. \sref{conclusion} will present what can be concluded and propose some future works.

%% file: chapters/related_work.tex
\section{Related Work}
\label{related_work}

Object detection has been one of the most studied fields in computer vision since the first use of \gls{cnn} \cite{lenet}. Since that, datasets have been released to improve object detection models. Among these released datasets, some label many classes, as COCO with 91 classes, and others are task-specific, with less than three classes. This section will present some of these datasets, as the COCO and some datasets used in other RoboCup leagues.

The most famous and used dataset for object detection is COCO \cite{dataset_coco}, released in 2014. This dataset contains 328.000 images, collected from Flickr, to avoid iconic-object images containing a single object centered in the image. Thus, the COCO dataset focus on non-iconic images, which means images with multiples categories in a diverse context. This strategy helps trained models to generalize objects instance, given the multiple contexts.

The classes used in the COCO dataset were chosen among 255 candidates given by children from 4 to 8 years old. The authors then voted on these categories based on how often each category occurred, and the most voted ones were selected, resulting in 80 classes. This dataset consists of 2.5 million instances, and as a result, each image averages 7.7 instances per image. It took 77.000 working hours to label all of these instances.

Moreover, in other RoboCup leagues, some datasets appear as good options. For instance, on \gls{msl}, there is an open object detection dataset \cite{dataset_msl}, which consists of 1456 images, divided into train and test using 70/30 proportion. This dataset uses images taken from the robot camera and images taken from outside of the field from different competitions to increase the variety of the dataset. This dataset provides the annotations in Pascal VOC and YOLO format, although it has only one class, labeling robot instances.

The \gls{spl} has an open-source tool to create and share dataset for object detection \cite{imagetagger} that has several images labeling Robot, Ball, and Goalpost. Besides, teams have been regularly releasing their datasets, as, per instance, the SPQR dataset \cite{spl_dataset}, that labels the same three classes. The SPQR dataset contains 2411 images collected from various game conditions, as natural and artificial light.

Other \gls{spl} dataset focus only on detecting Ball, as \cite{ball_spl}. This dataset has 6564 images collected from RoboCup logs and the authors' laboratory, varying lighting conditions. The images have a fixed size of $640\times480$ pixels from static and moving balls, resulting in 5209 ball examples.

%% file: chapters/dataset.tex
\section{Dataset}
\label{dataset}

\subsection{Dataset Creation}

The first step to create a labeled dataset is to select images to be part of it. The proposed dataset's images come from three different sources to use images under different conditions and angles. 

The first set of images consists of 259 pictures taken outside of the \gls{ssl} field, obtained from public image repositories of league teams. This set contains a variety of robot models and images taken under various light conditions. The second set has 516 brand-new images taken for this dataset from a smartphone camera inside a university laboratory field. Furthermore, the remaining 156 images were collected similarly to the second set but came from the final configuration, a camera placed on the \gls{ssl} robot. The images from this last set came from videos, where it was used a frame rate of 10 \gls{fps} to avoid using similar images. The combination of those sets results in a dataset of 931 images.

After collecting the images, they were resized to a standard resolution of $224 \times 224$ pixels as used by \cite{mobilenet,mobilenetv2}. \fref{fig:dataset_sample} shows some labeled examples from this dataset, each column of this image has two examples of each set of images.

\input{figures/dataset/dataset_sample}

The next step of creating the proposed dataset was to add labels to the objects in images. In the proposed dataset were defined three objects class to label: Robot, Ball, and Goal. These classes are the distinctive and relevant ones to detect in an \gls{ssl} game. Each image on this dataset can contain multiple labels, including none of it in the image. LabelImg \cite{labelimg} was used to label the images. This tool outputs squared detection in Pascal VOC and YOLO format.

After labeling the images, they were randomly divided into train and test set using the 70/30 proportion as in other robotics' object detection works \cite{dataset_msl}, \tref{tab:dataset_size} shows the final result of this division. This creation process took 160 working hours, most of them manually adding labels to each image. The proposed dataset is fully available on the author's GitHub \footnote{\url{https://github.com/bebetocf/ssl-dataset}}.

\input{tables/dataset/size}

\subsection{Dataset Statistics}

The proposed dataset's main objective is to detect objects in distinct game situations, so it is important to have multiple instances in each image. \fref{fig:instance_size} shows the number of instances per image. It is possible to see that most images have more than one instance, so there are more class instances than images. The dataset has 4182 instances, averaging 4.5 instances per image, which helps mitigate the low number of images.

\input{figures/dataset/instance_size}

\tref{tab:dataset_div} shows the instance division on the proposed dataset. The Goal class has fewer examples than Robot and Ball classes because not all images have a Goal instance, and when it appears, there is only one instance per image. Besides, the Goal instances are characteristics since they are very similar to each other. As COCO \cite{dataset_coco}, some datasets also have some imbalanced class, and some techniques could be used to balance a dataset \cite{oksuz2020imbalance}.

\input{tables/dataset/division}

The proposed dataset classifies the objects by their area, Small, Medium, and Large, similarly to COCO \cite{dataset_coco}. Small objects have less than $32\times32~(1024)$ pixels and represent 2919 instances. Medium objects are 1225 instances and have an area between $32\times32~(1024)$ pixels and $96\times96~(9216)$. Moreover, Large objects are bigger than $96\times96~(9216)$ pixels, representing 38 instances. Most objects concentrate in the Small area class, approximately 70\% of all objects, due to the low-resolution images.

\fref{fig:dataset_hist} shows each instance's division by the class and their area size. It is possible to see that almost all Ball instances are Small due to their actual size. More than half of the Robots' examples are Small due to images from the first set taken from outside the field, where robots are far from the camera. Furthermore, most of the Goals' samples are in Medium class.

\input{figures/dataset/size_histogram}

%% file: figures/dataset/dataset_sample.tex
\begin{figure}
\centering
\includegraphics[width=0.6\textwidth]{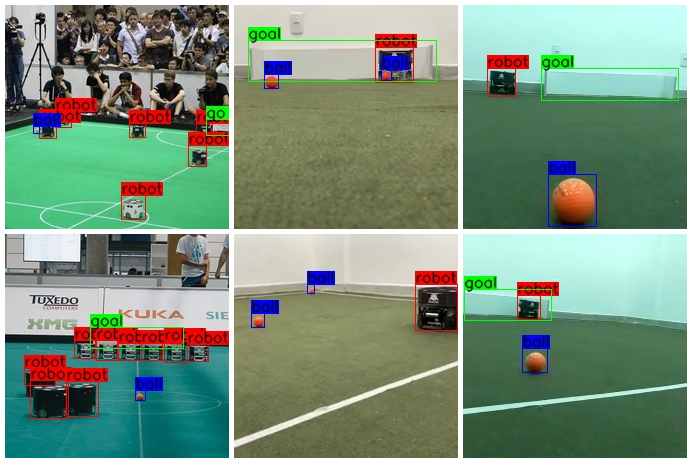}
\caption{Sample images from the dataset, showing ground-truth detection. The leftmost column has images from public \acrshort{ssl} images, the middle column has images collected from a smartphone inside the field, and the rightmost column has images taken in a camera placed in the robot.}
\label{fig:dataset_sample}
\end{figure} 

%% file: tables/dataset/size.tex
\begin{table}
\centering
\caption{Number of images divided into train and test set.}
\label{tab:dataset_size}
\begin{tabular}{p{0.2\textwidth}>{\centering}p{0.30\textwidth}}
\hline
&\textbf{Number of images} \cr
\hline
\textbf{Train}                          & 651 (69,92\%) \cr
\textbf{Test}                           & 280 (30,08\%) \cr \hline
\textbf{Dataset size}                          & 931 \cr 
\hline
\end{tabular}
\end{table}

%% file: figures/dataset/instance_size.tex
\begin{figure}
\centering
\includegraphics[width=0.6\textwidth]{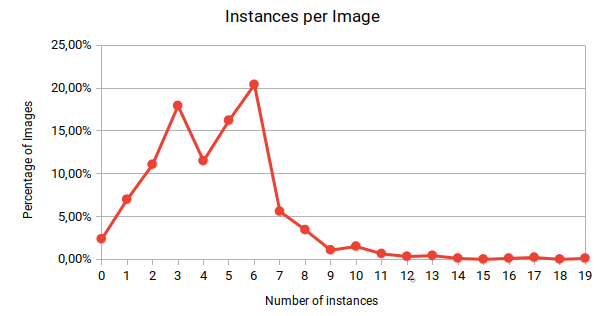}
\caption{Number of instances per image.}
\label{fig:instance_size}
\end{figure} 

%% file: tables/dataset/division.tex
\begin{table}
\centering
\caption{Number of instances of each class in the dataset.}
\label{tab:dataset_div}
\begin{tabular}{p{0.35\textwidth}>{\centering}p{0.30\textwidth}}
\hline
\textbf{Object Class} &\textbf{Instances per class} \cr
\hline
\textbf{Robot}                          & 1886 \cr
\textbf{Ball}                           & 1711 \cr
\textbf{Goal}                           & 585 \cr 
\hline
\textbf{Number of instances}                           & 4182 \cr 
\hline
\end{tabular}
\end{table}

%% file: figures/dataset/size_histogram.tex
\begin{figure}
\centering
\includegraphics[width=0.7\textwidth]{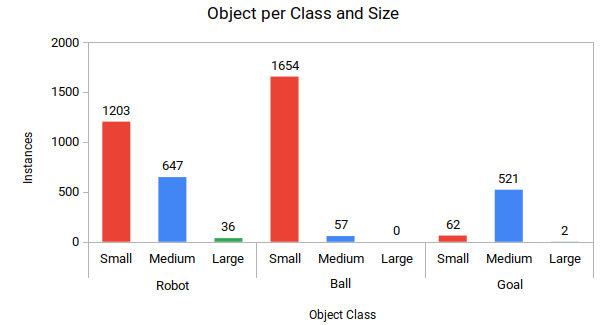}
\caption{Instance division per class and size. Small objects have less than $32 \times 32 ~ (1024)$ pixels, medium objects have a size between $32 \times 32 ~ (1024)$ pixels and $96 \times 96 ~ (9216)$ pixels, and large objects are bigger than $96 \times 96 ~ (9216)$ pixels.}
\label{fig:dataset_hist}
\end{figure} 

%% file: chapters/evaluation.tex
\section{Evaluation Methodology}
\label{evaluation}

\input{chapters/environment}
\input{chapters/models}

%% file: chapters/environment.tex
\subsection{Environment}
\label{environment}

One of the main drawbacks of using a \gls{cnn} is the requirement of a \gls{gpu} to infer in a frequency to use in a real-time detection \cite{gpucpu}. Besides, \gls{gpu} are too big to use in an \gls{ssl} robot, and they have a power consumption too high for the battery that fits in one of these robots. However, improvements in \gls{cnn} inference time in an embedded system make this method an excellent option to solve this technical challenge.

The primary constraint to this work is the environment delimitation due to the league's restrictions \cite{rules_ssl}. The robot used to test was a modified version on the RobôCIn v2020 \cite{tdp2020robocin}. As the technical challenge does not have a height restriction, another floor was added to the robot to fit additional hardware. All the modifications should have low power consumption, as the robot uses a LiPo 2200mah 4S 35C battery. This battery is enough to supply four brushless motors of 50W each and all the other robot's necessities.

A Raspberry Pi 4 Model B, a Google Coral Edge TPU accelerator, and a camera module were added to the robot, composing the vision system. The camera can capture images up to 90 \gls{fps} in a resolution of $640 \times 480$ pixels. These new components aim to tackle the lack of computational power in the main microcontroller, an STM32F767ZI. The power consumption of a Raspberry Pi 4 with the camera module is up to 7.5W, and the Google Coral is 4.5W, which fits the power supply of the robot's battery.

The vision system's inclusion is desirable to avoid modifying the architecture and data flow on the current robot. In the current robot, the microcontroller controls the motors to operate at the desired speeds. In the new system, the Raspberry Pi receives the camera's captured frames and uses them to input the inference model running on the Google Coral Edge TPU. After the inference, the model outputs the detected objects to the Raspberry Pi, which computes where the robot should go, and sends this position to the microcontroller. \fref{fig:system} shows the new system architecture of the robot.

\input{figures/environment/system_overview}

%% file: figures/environment/system_overview.tex
\begin{figure}[ht]
\centering
\includegraphics[width=0.9\textwidth]{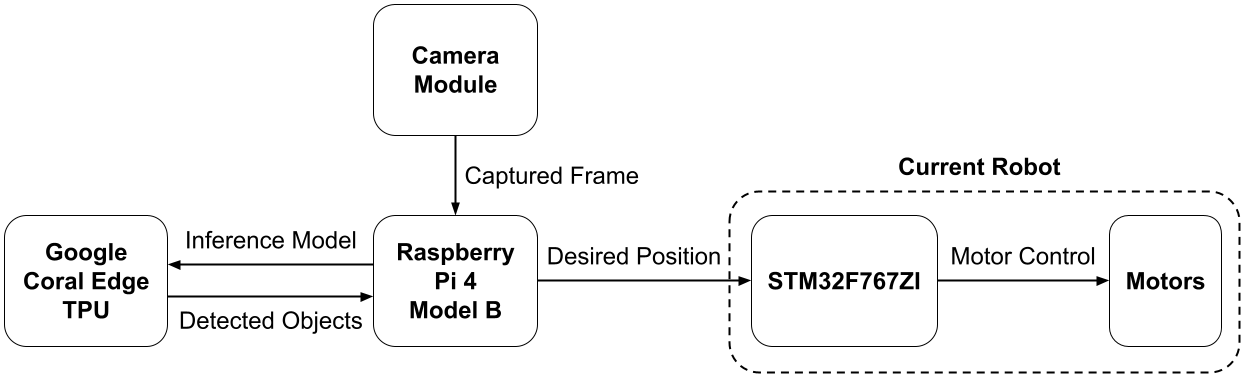}
\caption{Modifications on the architecture and data flow of the new robot.}
\label{fig:system}
\end{figure} 

%% file: chapters/models.tex
\subsection{Models and Experiments}
\label{models}

The pipeline to train, run and evaluate any model follows the same standards to each approach. Transfer Learning was used due to the proposed dataset size and the system restrictions, which speeds up training and takes advantage of low-level learned features \cite{transfer}. This technique uses a model pre-trained with another dataset to them train it with the proposed dataset. The proposed dataset was evaluated using MobileNet SSD v1 \cite{mobilenet}, MobileNet SSD v2 \cite{mobilenetv2}, MobileDet \cite{mobiledet} and YOLO v4 Tiny \cite{yolotiny}, which are state-of-the-art object detection models.

TensorFlow Object Detection API \cite{tf_object_api} was used to train MobileNets' and MobileDet's models. These models' train was improved using data augmentation techniques as Horizontal Flip, Image Crop, Image Scale, Brightness Adjustment, Contrast Adjustment, Saturation Adjustment, and Black Patches.

YOLO v4 tiny is a shallow version of the YOLO v4 \cite{yolov4}, designed to run in an embedded system. It already uses CutMix, Mosaic, Class Label Smoothing, and Self-Adversarial Training, so this architecture does not need any extra data augmentation technique. Furthermore, due to limitations of the portability process for Google Coral Edge TPU, the YOLO v4 tiny uses ReLU rather than Leaky ReLU as activation functions. 

These models were optimized using Integer Quantization, which increases the inference speed while maintaining network precision \cite{quantization}. This method consists of converting the network weight from floating-point numbers to integer values. After training, the models were quantized and converted to a TensorFlow Lite compatible model, which is required to compile the model to run on a Google Coral Edge TPU accelerator.

\subsection{Running and Evaluating}
\label{running}

The primary constraint to use a trained model on an \gls{ssl} robot is inferring in real-time. A model has to run in at least 24 \gls{fps} to be considered a real-time inference. This frame rate is acceptable with the league's objects' speed since a ball, the fastest object in the field, with a maximum speed of $6.5 m/s$ \cite{rules_ssl}, would move only $27cm$ between inferences.

The models were evaluated using the metrics as the COCO dataset, which are \gls{ap} and \gls{ar}. In those metrics, to determine if a detected object is a true positive or a false positive, it has to define a \gls{iou} threshold to consider a correct prediction. The \gls{ap} and \gls{ar} metrics uses the mean of ten \gls{iou} threshold values from $0.5$ to $0.95$ with a step of $0.05$. Besides, these metrics are presented by each object's size. The evaluation is made using an open-source tool \cite{metrics_tool}, that given ground-truth labels and predictions, output the COCO metrics to compare each model's result.

%% file: chapters/results.tex
\section{Results}
\label{results}

\tref{tab:results_coco_ap} shows the \gls{ap} for the four models separated by \gls{iou} threshold and object area. \gls{ap} for Medium and Large objects shows how powerful these models can be in less challenging scenarios, where the object is much closer to the robot. However, the results for Small objects are worse than for Medium and Large objects, which indicates a high false-positive rate. This error occurs due to the low information on objects of Small size.

\input{tables/results/coco_ap}

From the \gls{ap} perspective, the MobileNet SSD v1 had the best result overall and for Large objects. The \gls{ap} for Large objects on the YOLO v4 tiny model was worse than for Medium objects, which is a peculiar behavior since the other models achieve better \gls{ap} when detecting Large objects. This result can indicate that YOLO v4 tiny needs more labeled data with Large size, as there are only 38 objects with this size on the proposed dataset.

\tref{tab:results_coco_ar} shows the \gls{ar} results separated by maximum detection per image and detected object size. A high \gls{ar} is important for Robot and Goal classes, as the robot uses it to avoid colliding with another robot when navigating and helps the robot identify the Goal faster. The Robot and Goal classes' represents all of the Large objects and $95\%$ of Medium objects, as shown in \fref{fig:dataset_hist}.

\input{tables/results/coco_ar}

The obtained result of \gls{ar} for Medium and Large objects sizes shows a high detection rate, with the MobileNet SSD v1 as the best \gls{ar} results overall. It was also observed that YOLO v4 tiny had worse results for Large objects, which supports the necessity of more samples of Large objects for this model.

\tref{tab:results_fps} shows each model's inference frequency, where the MobileNet SSD v1 had the best \gls{fps} overall, but MobileNet SSD v2 and MobileDet had a rate that fits the requirement of at least 24 \gls{fps}. However, the YOLO v4 Tiny had a bad result with only 10 \gls{fps}, which is caused by the lack of architectural optimizations on the network compared with the other evaluated models. This shortage of optimization results in a bad mapping to the Google Coral.

\input{tables/results/fps}

\fref{fig:precision_recall} shows the Precision-Recall curve for each model, separated by class, and using a \gls{iou} threshold of $0.5$. The Ball class is the most difficult class to detect due to the object size. The MobileNet SSD v2 in this \gls{iou} had a good result in all the three classes but has a smaller recall in the Goal class. The YOLO v4 tiny had a good precision when detecting all classes but could detect only $20\%$ of the Ball in the test set. This result explains why the \gls{ap}$_{50}$ for this model is lower than MobileNet SSD v2 since it is calculated averaging precision across recalls values from 0 to 1. This figure also shows that the fewer examples in the Goal class were not a problem since the Precision x Recall curve for this class is very similar to the other classes.

\input{figures/results/precision_recall_all}

%% file: tables/results/coco_ap.tex
\begin{table}
\centering
\caption{\acrfull{ap} for each model by \acrfull{iou} threshold, in \acrshort{ap}$_{50}$ the threshold used is 0.5 and \acrshort{ap}$_{75}$ is 0.75, and detected object area, where \acrshort{ap}$_{S}$, \acrshort{ap}$_{M}$, and \acrshort{ap}$_{L}$ stands for the result by each detection size, Small, Medium or Large.}
\label{tab:results_coco_ap}
\begin{tabular}{p{0.26\textwidth}>{\centering}p{0.11\textwidth}>{\centering}p{0.11\textwidth}>{\centering}p{0.11\textwidth}>{\centering}p{0.11\textwidth}>{\centering}p{0.11\textwidth}>{\centering}p{0.11\textwidth}}
\hline
\textbf{Method}&\textbf{\acrshort{ap}}&\textbf{\acrshort{ap}$_{50}$}&\textbf{\acrshort{ap}$_{75}$}&\textbf{\acrshort{ap}$_{S}$}&\textbf{\acrshort{ap}$_{M}$}&\textbf{\acrshort{ap}$_{L}$} \cr
\hline
\textbf{MobileNet SSD v1} & \textbf{44.88\%} &         68.81\%  &         47.51\%  &         26.83\% &          68.54\%  & \textbf{89.62\%} \cr
\textbf{MobileNet SSD v2} &         43.42\%  & \textbf{74.41\%} &         44.83\%  &         23.06\% &          62.55\%  &         82.93\% \cr
\textbf{MobileDet}    &         35.96\%  &         64.95\%  &         36.62\%  &         16.63\% &          60.48\%  &         82.97\%  \cr
\textbf{YOLOv4 Tiny}  &         42.17\%  &         62.24\%  & \textbf{54.09\%} & \textbf{27.34\%} & \textbf{69.05\%} &         58.84\% \cr
\hline
\end{tabular}
\end{table}

%% file: tables/results/coco_ar.tex
\begin{table}
\centering
\caption{\acrfull{ar} for each model by maximum detection per image,  \acrshort{ar}$_{1}$ for at most 1 object per image and \acrshort{ar}$_{10}$ for 10 objects per image, and detected object area, where \acrshort{ar}$_{S}$, \acrshort{ar}$_{M}$, and \acrshort{ar}$_{L}$ stands for the result by each detection size, Small, Medium or Large.}
\label{tab:results_coco_ar}
\begin{tabular}{p{0.25\textwidth}>{\centering}p{0.13\textwidth}>{\centering}p{0.13\textwidth}>{\centering}p{0.13\textwidth}>{\centering}p{0.13\textwidth}>{\centering}p{0.13\textwidth}}
\hline
\textbf{Method}&\textbf{\acrshort{ar}$_{1}$}&\textbf{\acrshort{ar}$_{10}$}&\textbf{\acrshort{ar}$_{S}$}&\textbf{\acrshort{ar}$_{M}$}&\textbf{\acrshort{ar}$_{L}$} \cr
\hline
\textbf{MobileNet SSD v1} & \textbf{37.75\%} & \textbf{62.87\%} & \textbf{40.62\%} & \textbf{82.18\%} & \textbf{91.00\%} \cr
\textbf{MobileNet SSD v2} &         34.76\%  &         50.60\%  &         29.28\%  &         66.21\%  &         87.00\%  \cr
\textbf{MobileDet}    &         30.62\%  &         43.66\%  &         22.96\%  &         65.41\%  &         85.00\%  \cr
\textbf{YOLOv4 Tiny}  &         36.72\%  &         45.36\%  &         30.41\%  &         74.16\%  &         64.00\%  \cr
\hline
\end{tabular}
\end{table}

%% file: tables/results/fps.tex
\begin{table}
\centering
\caption{Mean inference frequency in \acrfull{fps} for each tested model.}
\label{tab:results_fps}
\begin{tabular}{p{0.3\textwidth}>{\centering}p{0.1\textwidth}}
\hline
\textbf{Method}&\textbf{FPS} \cr
\hline
\textbf{MobileNet SSD v1} & $\textbf{94}$ \cr
\textbf{MobileNet SSD v2} & $78$ \cr
\textbf{MobileDet}    & $87$ \cr
\textbf{YOLOv4 Tiny}  & $10$ \cr
\hline
\end{tabular}
\end{table}

%% file: figures/results/precision_recall_all.tex
\begin{figure}[ht]
\centering
\includegraphics[width=0.7\textwidth]{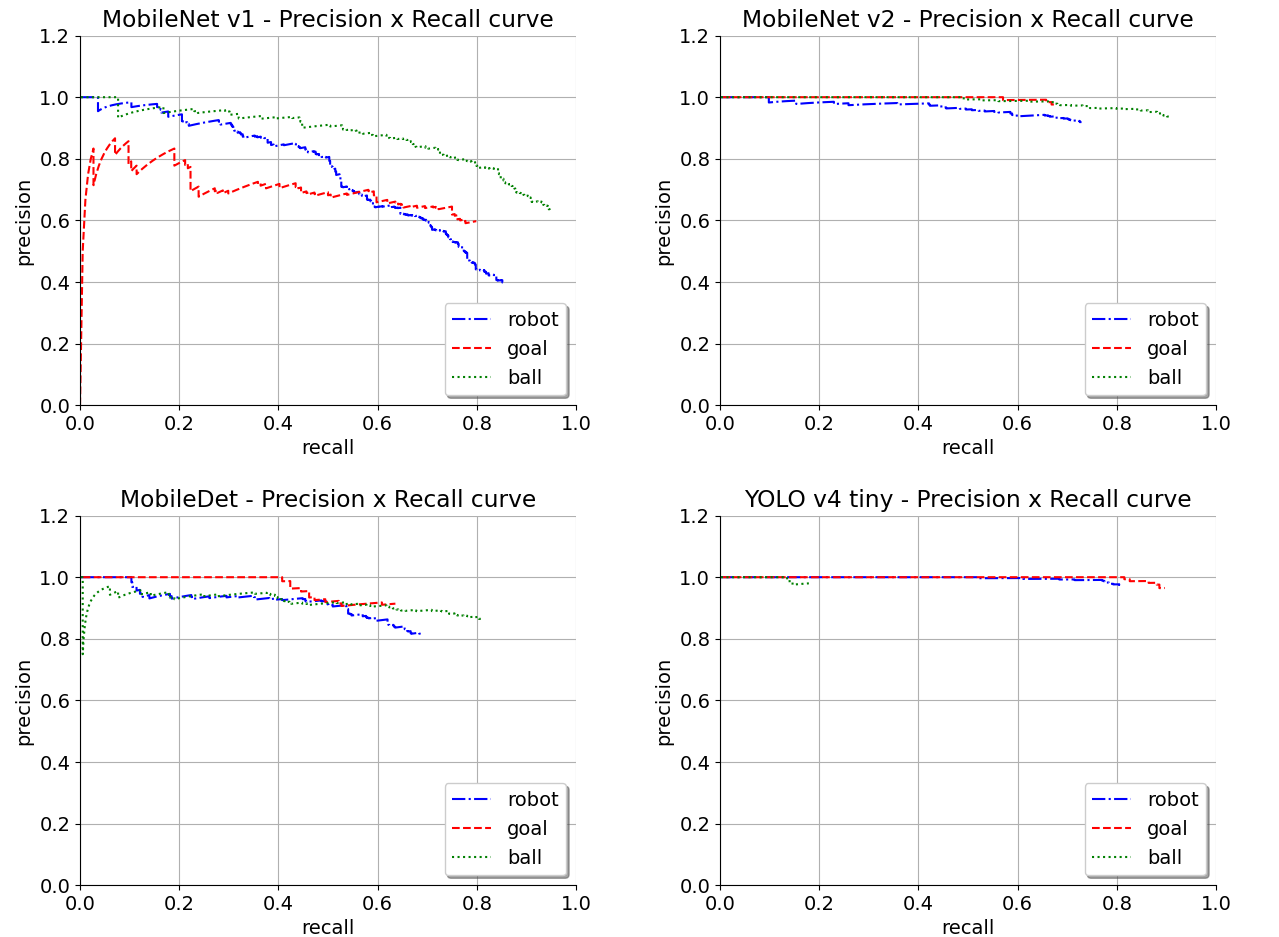}
\caption{Precision x Recall graphs for trained models, using \gls{iou} = 0.5. When the line disappears, it means that the value of precision is zero.}
\label{fig:precision_recall}
\end{figure} 

%% file: chapters/conclusion.tex
\section{Conclusion}
\label{conclusion}

This paper presented an open-source labeled dataset and a benchmark for object detection for \gls{ssl}. The proposed dataset guarantees variety with images extracted from different sources under distinct lighting conditions and camera configurations. The labeled objects are Robot, Ball, and Goal, which are the essential objects found during an \gls{ssl} game. The dataset's images can contain multiple instances of these objects, including no objects per image.

It was also presented a pipeline to train a \gls{cnn} and deploy it on an embedded device with limited computational power. The results show that \gls{cnn} are robust to variable light conditions and can also detect robots with different structures. This result contrasts with using color segmentation with scan lines. Color segmentation can be easily disturbed by these circumstances since it needs fine-tuning parameters that rely on image saturation and brightness.

The presented dataset has a similar size as other datasets used on other RoboCup leagues. However, it is smaller than general propose object detection datasets. So, data augmentation techniques were applied to increase diversity and model generalization. A future improvement to the dataset is to add images from game situations and different field configurations. Besides, increasing the number of distinctive robots' instances and the number of Large images will boost the dataset robustness.

This paper uses the proposed dataset to evaluate \gls{ap}, \gls{ar}, and \gls{fps} of four different \gls{cnn} models on constrained hardware. Furthermore, this paper highlights the importance of model architectural optimizations. Future works will analyze other models and modifications to hyperparameters, as input size, to enhance Small object detection. Further work will also analyze other techniques, such as tracking, to continuously detect objects in multiple frames in a real game environment.

%% file: default_content/acknowledgements.tex
\subsubsection{Acknowledgements}

The authors would like to acknowledge the RoboCIn's team and Centro de Informática - UFPE for all the research support. The first and second authors were also funded by the Conselho Nacional de Desenvolvimento Científico e Tecnológico (CNPq). The authors appreciate all the \gls{ssl}'s teams for the open-source images from past competitions.